\documentclass[runningheads]{llncs}

\usepackage{graphicx}
\usepackage{amsmath,amssymb}  
\usepackage{multirow}
\usepackage[utf8]{inputenc} 
\usepackage[T1]{fontenc}    
\usepackage{bbm, dsfont}

\usepackage{url}            
\usepackage{booktabs}       
\usepackage{amsfonts}       
\usepackage{nicefrac}       

\usepackage{siunitx}
\newcommand{\figref}[1]{Fig.~\ref{#1}}

\newcommand{\tabref}[1]{Tab.~\ref{#1}}

\newcommand{\R}[1]{\textcolor{red}{#1}}

\newcommand{\B}[1]{\textcolor{blue}{#1}}

\usepackage{authblk}

\usepackage{color}

\usepackage[pagebackref=true,breaklinks=true,letterpaper=true,colorlinks,bookmarks=false]{hyperref}

\begin{document}
\pagestyle{headings}
\mainmatter

\def\ACCV20SubNumber{***}  

\title{IAFA: Instance-aware Feature Aggregation for 3D Object Detection from a Single Image} 
\titlerunning{IAFA for 3D Object Detection from a Single Image}
%
\author{Dingfu Zhou\inst{1,2}, Xibin Song\inst{1,2}, Yuchao Dai\inst{3}, Junbo Yin \inst{1,4}, Feixiang Lu \inst{1,2}, Jin Fang \inst{1,2}, Miao Liao \inst{1} and Liangjun Zhang \inst{1}}
\authorrunning{Zhou, D., et al.}
\institute{Baidu Research \and National Engineering Laboratory of Deep Learning Technology and Application, Beijing, China \and Northwestern Polytechnical University, Xi'an, China \and Beijing Institute of Technology, Beijing, China }

\maketitle

\begin{abstract}
3D object detection from a single image is an important task in Autonomous Driving (AD), where various approaches have been proposed. However, the task is intrinsically ambiguous and challenging as single image depth estimation is already an ill-posed problem. In this paper, we propose an instance-aware approach to aggregate useful information for improving the accuracy of 3D object detection with the following contributions. First, an instance-aware feature aggregation (IAFA) module is proposed to collect local and global features for 3D bounding boxes regression. Second, we empirically find that the spatial attention module can be well learned by taking coarse-level instance annotations as a supervision signal. The proposed module has significantly boosted the performance of the baseline method on both 3D detection and 2D bird-eye's view of vehicle detection among all three categories. Third, our proposed method outperforms all single image-based approaches (even these methods trained with depth as auxiliary inputs) and achieves state-of-the-art 3D detection performance on the KITTI benchmark.
\end{abstract}

\section{Introduction}
Accurate perception of surrounding environment is particularly important for Autonomous Driving \cite{geiger2012we}, \cite{wang2019apolloscape},  and robot systems. In AD pipeline, the perception 3D positions and orientation of surrounding obstacles (e.g., vehicle, pedestrian, and cyclist) are essential for the downstream navigation and control modules. 3D object detection with depth sensors (e.g., RGBD camera, LiDAR) is relatively easy and has been well studied recently. Especially, with the development of deep learning techniques in 3D point cloud, a wide variety of 3D object detectors have sprung up including point-based methods \cite{qi2018frustum}, \cite{shi2019pointrcnn},  voxel-based methods \cite{zhou2018voxelnet}, \cite{lang2018pointpillars} \cite{yin2020lidar} and hybrid-point-voxel-based methods \cite{yang2019std}, \cite{shi2020pv}. 

Although depth sensors have been widely used in different scenarios, their drawbacks are also obvious: expensive prices, high-energy consumption, and less structure information. Recently, 3D object detection from passive sensors such as monocular or stereo cameras has attracted many researchers' attention and some of them achieved impressive detection performance. Compared with the active sensors, the most significant bottleneck of 2D image-based approaches \cite{zhou2017moving} is how to recover the depth of these obstacles. For stereo rig, the depth (or disparity) map can be recovered by traditional geometric matching \cite{hernandez2008multiview} or learned by deep neural networks \cite{luo2016efficient}. By using traditional geometric techniques, it's really difficult to estimate the depth map from a single image without any prior information while this problem has been partly solved with deep learning based methods, \cite{godard2019digging}. With the estimated depth map, the 3D point cloud (pseudo LiDAR point cloud) can be easily reconstructed via pre-calibrated intrinsic (or extrinsic) camera parameters. Any 3D detectors designed for LiDAR point cloud can be use directly on pseudo LiDAR point cloud \cite{qian2020end}, \cite{wang2019pseudo}, \cite{weng2019monocular} and \cite{vianney2019refinedmpl}. Furthermore, in \cite{qian2020end}, the depth estimation and 3D object detection network has been integrated together in an end-to-end manner.

 \begin{figure}[t!]
 	\centering
 	\includegraphics[width=0.99\textwidth]{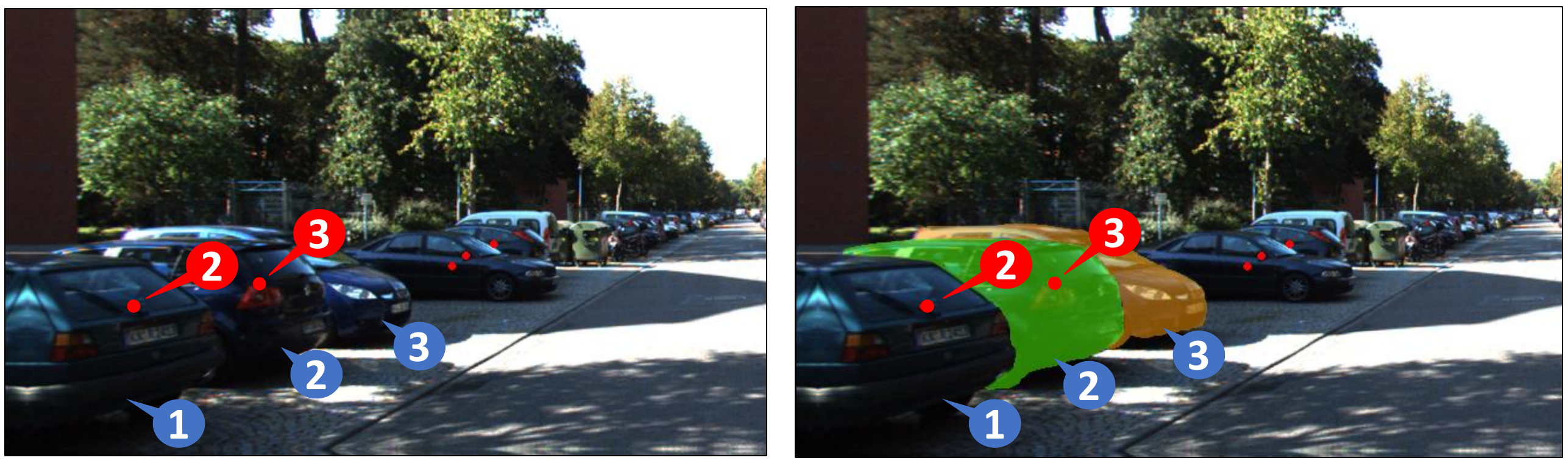}
 	\caption{An example of center-point-based object representation. The red points in the left sub-image are the projected 3D center of the objects on the 2D image. The numbers in the blue text box represent the ``ID'' of vehicles. The red points are the projected 3D vehicle centers onto the 2D image plane. The number in the red text boxes represents which vehicle that the center point belongs to. }
 	\label{fig:center_point_based_object_representation}
 \end{figure}

Recently, center-based (anchor-free) frameworks become popular for 2D object detection. Two representative frameworks are ``CenterNet'' \cite{zhou2019objects} and ``FCOS'' \cite{tian2019fcos}. Inspired by these 2D detectors, some advanced anchor-free 3D object detectors have been proposed such as ``SMOKE''\cite{liu2020smoke} and ``RTM3D'' \cite{li2020rtm3d}. In the center-based frameworks, an object is represented as a center point and object detection has been transferred as a problem of point classification and its corresponding attributes (e.g., size, offsets, depth, etc.) regression.  

Although the center-based representation is very compact and effective, it also has some drawbacks. In the left sub-figure of \figref{fig:center_point_based_object_representation}, the vehicles are represented with center points, where the red points are the projected 3D object centers on the 2D image plane. The white numbers in the blue text boxes represent the ``ID'' of the vehicles and the white numbers in the red text boxes are the vehicle ``ID''s that these points belong to. From this image, we can easily find that the projected 3D centers of vehicle ``2'' is on the surface of vehicle ``1''. Similarly, the projected 3D centers of vehicle ``3'' is on the surface of vehicle ``2''. Particularly, this kind of misalignment commonly happens in the AD scenario in the case of occlusion. Taking vehicle ``2'' as an example, its projected 3D center is on the surface of vehicle ``1'' and most of its surrounding pixels are from vehicle ``1''. During the training, the network may be confused about which pixels (or features) should be used for this center classification and its attributes regression. This problem becomes much more serious for the depth regression because the real depth of the 2D center point (on the surface of vehicle ``1'') is totally different from its ground truth value-the the depth of its 3D Bounding Box's (BBox's) center.

In order to well handle this kind of misalignment or to alleviate this kind of confusion during the network learning process, we propose to learn an additional attention map for each center point during training and explicitly tell the network that which pixels belong to this object and they should contribute more for the center classification and attributes regression. Intuitively, the learning of the attention map can be guided by the instance mask of the object. By adding this kind of attention map estimation, we can achieve the following advantages: first of all, the occluded objects, attention map can guide the network to use the features on the corresponding objects and suppress these features that belong to the other object; second, for these visible objects, the proposed module is also able to aggregate the features from different locations to help the object detection task. The contributions of our work can be summarized as follows: 
\begin{enumerate}
\item First, we propose a novel deep learning branch termed as Instance-Aware Features Aggregation (IAFA) to collect all the pixels belong to the same object for contributing to the object detection task. Specifically, the proposed branch explicitly learns an attention map to automatically aggregate useful information for each object.   
\item Second, we empirically find that the coarse instance annotations from other instance segmentation networks can provide good supervision to generate the features aggregation attention maps. 
\item The experimental results on the public benchmark show that the performance of the baseline can be significantly improved by adding the proposed branch. In addition, the boosted framework outperforms all other monocular-based 3D object detectors among all the three categories (``Easy'', ``Moderate'' and ``Hard'').
\end{enumerate}

\section{Related Work}

\textbf{LiDAR-based 3D Detection:} 3D object detection in traffic scenario becomes popular with the development of range sensor and the AD techniques. Inspired by 2D detection, earlier 3D object detectors project point cloud into 2D (e.g., bird-eye-view \cite{chen2016monocular} or front-view \cite{wu2018squeezeseg}) to obtain the 2D object bounding boxes first and then re-project them into 3D. With the development of 3D convolution techniques, the recently proposed approaches can be generally divided into two categories: volumetric convolution-based methods and points-based methods. Voxel-net \cite{zhou2018voxelnet} and PointNet \cite{qi2017pointnet} are two pioneers for these methods, respectively. How to balance the GPU memory consumption and the voxel's resolution is one bottleneck of voxel-based approach. At the beginning, the voxel resolution is relative large as $0.4 \text{m} \times 0.2 \text{m} \times 0.2 \text{m} $ due the limitation of GPU memory. Now this issue has been almost solved due to the development of GPU hardware and some sparse convolution techniques, e.g., SECOND \cite{yan2018second} and PointPillars \cite{lang2018pointpillars}. At the same time, points-based methods\cite{zhou2020joint} also have been well explored and achieved good performance on the public benchmarks. 
 
\noindent\textbf{Camera-based 3D object Detection:}
due to the cheaper price and less power consumption,  many different approaches have been proposed recently for 3D object detection from camera sensor. 
 A simple but effective idea is to reconstruct the 3D information of the environment first and then any point cloud-based detectors can be employed to detect objects from the reconstructed point clouds (which is also called ``Pseudo-LiDAR'') directly. For depth estimation (or 3D reconstruction), either classical geometric-based approaches or deep-learning based approaches can be used. Based on this idea, many approaches have been proposed for either monocular \cite{weng2019monocular} or stereo cameras \cite{qian2020end}, \cite{wang2019pseudo}, \cite{ma2019accurate}, \cite{vianney2019refinedmpl}. Rather than transforming the depth map into point clouds, many approaches propose using the depth estimation map directly in the framework to enhance the 3D object detection. In M3D-RPN \cite{brazil2019m3d} and \cite{ding2020learning}, the pre-estimated depth map has been used to guide the 2D convolution, which is called as ``Depth-Aware Convolution''. In addition, in order to well benefit the prior knowledge, some approaches are also proposed to integrate the shape information into the object detection task via sparse key-points \cite{song2019apollocar3d} or dense 2D and 3D mapping .  

\noindent\textbf{Direct Regression-based Methods:}
although these methods mentioned above achieved impressive results, they all need auxiliary information to aid the object detection, such as ``Depth Map'' or ``Pseudo Point Cloud''. Other approaches are direct regression-based methods. Similar to the 2D detectors, the direct regression based methods can be roughly divided into anchor-based or anchor-free methods. Anchor-based methods such as \cite{li2019gs3d}, \cite{liu2019deep} \cite{jorgensen2019monocular}, \cite{chen2020monopair} need to detect 2D object bounding boxes first and then ROI align technique is used to crop the related information in both original image domain or extracted feature domain for corresponded 3D attributes regression. Inspired by the development of center-point-based (anchor free) methods in 2D object detection \cite{zhou2019objects}, \cite{tian2019fcos} and instance segmentation \cite{lee2020centermask}, \cite{wang2020centermask}, some researchers have proposed center-point-based methods for 3D object detection task \cite{li2020rtm3d}, \cite{liu2020smoke} and \cite{chen2020monopair}. In \cite{zhou2019objects}, the object has been represented as a center point with associated attributes (e.g., object's size, category class, etc.). In addition, they extend this representation into 3D and achieve reasonable performance. Based on this framework, Liu et.al \cite{liu2020smoke} modify the baseline 3D detector by adding the group-normalization in backbone network and propose a multi-step disentangling approach for constructing the 3D bounding box. With these modifications, the training speed and detection performances have been significantly improved. Instead of representing the object as a single point, Li et.al., \cite{li2020rtm3d} propose to use nine points which are center point plus eight vertexes of the 3D bounding box. First, the network is designed to detect all the key-points and a post-processing step is required for solving the object pose as an optimization problem.    

\noindent\textbf{Attention-based Feature Aggregation:}
recently, attention-based feature aggregation strategies have proven their effectiveness in many areas, such as image super-resolution~\cite{liu2019image}, image translation~\cite{Tang_2019_CVPR}, \cite{8755856}, GAN based methods~\cite{zhang2019self}, semantic segmentation~\cite{fu2019dual}. According to previous work, the attention strategies can efficiently enhance extracted features in several manners, including: channel attention aggregation~\cite{song2020channel} and spatial attention based aggregation~\cite{liu2019image}, \cite{fu2019dual}. The channel attention based aggregation strategy aims to learn the weight of each channel of feature maps to aggregate the features in channel-level, while spatial attention based aggregation aims to learn the weight of each pixel in feature maps to aggregate the features in pixel-level.
\section{Definition and Baseline Method}
Before the introduction of the proposed approach, the 3D object detection problem and the baseline center-based framework will be discussed first. 
\subsection{Problem Definition}\label{subsec:problem_definition}
For easy understanding, the camera coordinate is set as the reference coordinate and all the objects are defined based on it. In deep-learning-based approaches, an object is generally represented as a rotated 3D BBox as 
\begin{equation}
    \textbf{c}, \textbf{d}, \textbf{r} = (c_x, c_y, c_z), (l, w, h), (r_x, r_y, r_z),
\end{equation}
in which $\textbf{c}$, $\textbf{d}$, $\textbf{r}$ represent the centroid, dimension and orientation of the BBox respectively. In AD scenario, the road surface that the objects lie on is almost flat locally, therefore the orientation parameters are reduced from three to one by keeping only the yaw angle $r_y$ around the Y-axis. In this case, the BBox is simply represented as $(c_x$, $c_y$, $c_z$, $l$, $w$, $h$, $r_y)$.

\subsection{Center-based 3D Object Detector}
Center-based (anchor-free) approaches have been widely employed for 2D object detection and instance segmentation recently. In these methods, an object is represented as a center with associated attributes (e.g., dimensions and center offsets) which are obtained with a classification and regression branches simultaneously. Based on the 2D centernet, Liu et al. \cite{liu2020smoke} modified and improved it for 3D object detection, where the object center is the projection of 3D BBox's centroid and the associated attributes are 3d dimensions, depth and object's orientation etc.
 \begin{figure}[h!]
 	\centering
 	\includegraphics[width=0.99\textwidth]{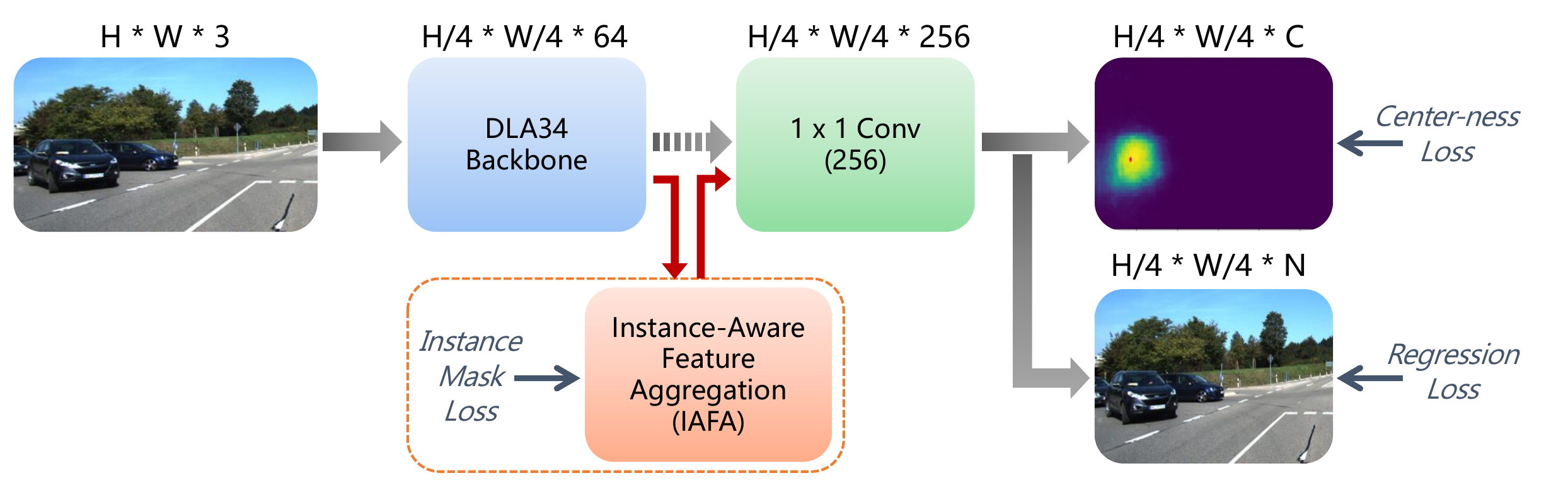}
 	\caption{A sketch description of the proposed Instance-Aware Feature Aggregation (IAFA) module integrated with the baseline 3D object detector. In which, the structure inside the dotted bordered rectangle is the proposed IAFA and ``C'' and ``N'' at the end of the frameworks are the number of ``categories'' and regression parameters respectively.}
 	\label{fig:FrameWork}
 \end{figure}

 A sketch of baseline 3D object detector is illustrated in \figref{fig:FrameWork}. By passing the backbone network (e.g., DLA34 \cite{yu2018deep}), a feature map $\mathbf{F}_\text{backbone}$ with the size of $\frac{\text{W}}{4} \times \frac{\text{H}}{4} \times 64$ is generated from the input image $\mathbf{I}$ ($\text{W} \times \text{H} \times 3$). After a specific $1 \times 1 \times 256$ convolution layer, two separate branches are designed for center classification and corresponding attributes regression. Due to anchor-free,  the classification and regressions are generated densely for all points of the feature map. In center classification branch, a point is classified as positive if its response is higher than a certain threshold. At the same time, its associated attributes can be obtained correspondingly according to its location index.  
 
 \textbf{3D BBox Recovery:} assuming that a point $(x_i, y_i)$ is classified as an object's center and its associated attributes usually includes $(x_\text{offset}, y_\text{offset})$, $depth$, $(l, w, h)$ and $(\text{sin} \theta, \text{cos} \theta)$, where $d$ is the depth of object, $(l, w, h)$ is 3D BBox's dimension. Similar to \cite{mousavian20173d}, $\theta$ is an alternative representation of $r_y$ for easy regression and $(x_\text{offset}, y_\text{offset})$ is estimated discretization residuals due to feature map downsampling operation. Based on the 2D center and its attributes, the 3D centroid $(c_x, c_y, c_z)$ of the object can be recovered via

\begin{equation}
[c_x, c_y, c_z ]^\text{T} = \mathbf{K}^{-1} * [ x_i + x_\text{offset}, y_i + y_\text{offset}, 1]^\text{T} * \text{depth} , 
\end{equation} 
where $\mathbf{K}$ is the camera intrinsic parameter. 

 \textbf{Loss Function:} 
 during training, for each ground truth center $p_k$ of class $j$, its corresponding low-resolution point $\hat{p}_j$ in the down-sampled feature map is computed first. To increase the positive sample ratio, all the ground truth centers are splat onto a heatmap  $\mathbf{h} \in [0, 1]$ with the size of $ \frac{\text{W}}{4} \times \frac{\text{H}}{4} \times C $ using a Gaussian kernel $h_{xyj} = \text{exp}(-\frac{(x-\hat{p}_x)^2 + (y - \hat{p}_y)^2}{2\sigma^{2}_{p}}$, where $\sigma_{p}$ is an object size-adaptive standard deviation. If two Gaussians of the same class overlap, the element-wise maximum is employed here. The training loss for center point classification branch is defined as 
\begin{equation}
   \text{L}_\text{center-ness}=\frac{-1}{M}\sum_{xyj}\left\{
\begin{array}{ll}
(1-\hat{h}_{xyj})^{\alpha} \text{log}(\hat{h}_{xyj})         & if \: h_{xy} = 1,\\
(1-h_{xyj})^{\beta}(\hat{h}_{xyj})^{\alpha} \text{log}(1-\hat{h}_{xy})         & \text{otherwise.}
\end{array} \right. 
\label{eq:center-ness-loss}
\end{equation} 
where $\alpha$ and $\beta$ are hyper-parameters of the focal loss \cite{lin2017focal} and $M$ is the number of all positive center points. 

Although the attributes regression is computed densely for each location in the feature map, the loss function is only defined sparsely on the ground truth centers. Usually, a general expression of regression loss is defined as

\begin{equation}
    \text{L}_\text{reg} = \frac{1}{N} \sum_{i=1}^{N}(\mathbbm{1}_{{p}_{i}} l_\text{reg}), \:  \mathbbm{1}_{{p}_{i}} = \left\{
\begin{array}{ll}
1    & if \: {p}_{i}  \: \text{is an object center},\\
0    & \text{otherwise.} 
\end{array} \right. 
\label{eq:regression-loss}
\end{equation}
where $l_\text{reg}$ is a general definition of regression loss which can be defined as $\text{L}_{1}$ or ${smooth-L}_{1}$ loss defined on the prediction directly, $corners$ loss \cite{liu2020smoke} on the vertex of the recovered 3D BBox, IoU loss \cite{zhou2019iou} on 3D BBoxes or disentangling detection loss \cite{simonelli2019disentangling} etc.

\section{Proposed Approach}

We propose the IAFA network to gather all the useful information related to a certain object for 3D pose regression. It generates a pixel-wise spatial attention map to aggregate all the features belongs to the objects together for contributing the center classification and its attribution regression. The proposed branch is a light-weight and plug-and-play module, which can be integrated into any one-stage based 3D object detection framework. 

\subsection{IAFA Module}
 The proposed IAFA branch is highlighted with dotted box in \figref{fig:FrameWork}, which aims at collecting all the useful information (e.g., related to a certain object) together to help 3D object detection task. Specifically, for a feature map $\mathbf{F}^{s}$ in a certain level, with the size of $\text{W}^s \times \text{H}^s \times \text{C}^s$, the IAFA module will generate a high-dimension matrix $\mathbf{G}$ with the size of $\text{W}^s \times \text{H}^s \times \text{D}$, where $\text{D} = \text{W}^s \times \text{H}^s$. For a certain location $(i, j)$ of $\mathbf{G}$, the vector $\mathbf{G}_{ij} \in \mathbb{R}^{1 \times \text{D}}$ encodes a dense relationship map of the target point $p (i, j)$ with all the other locations (including itself). Intuitively, these pixels belonging to the same object should have closer relationship than those pixels that don't belong to the object and therefore they should give more contribution for the 3D object detection task. 
 
 To achieve this purpose, we propose to use the corresponding object instance mask as a supervised signal for learning this attention map $\mathbf{G}$. For efficient computation, this supervision signal is only sparsely added to object's centers. Some learned attention vectors $\mathbf{G} (i, j)$ (reshaped as images with the size of $\text{W}^s \times \text{H}^s$ for easy understanding) are displayed in \figref{fig:Example_of_AttentionMap}, in which three maps correspond to three objects' centers. 
\begin{figure}[t!]
 	\centering
 	\includegraphics[width=1.0\textwidth]{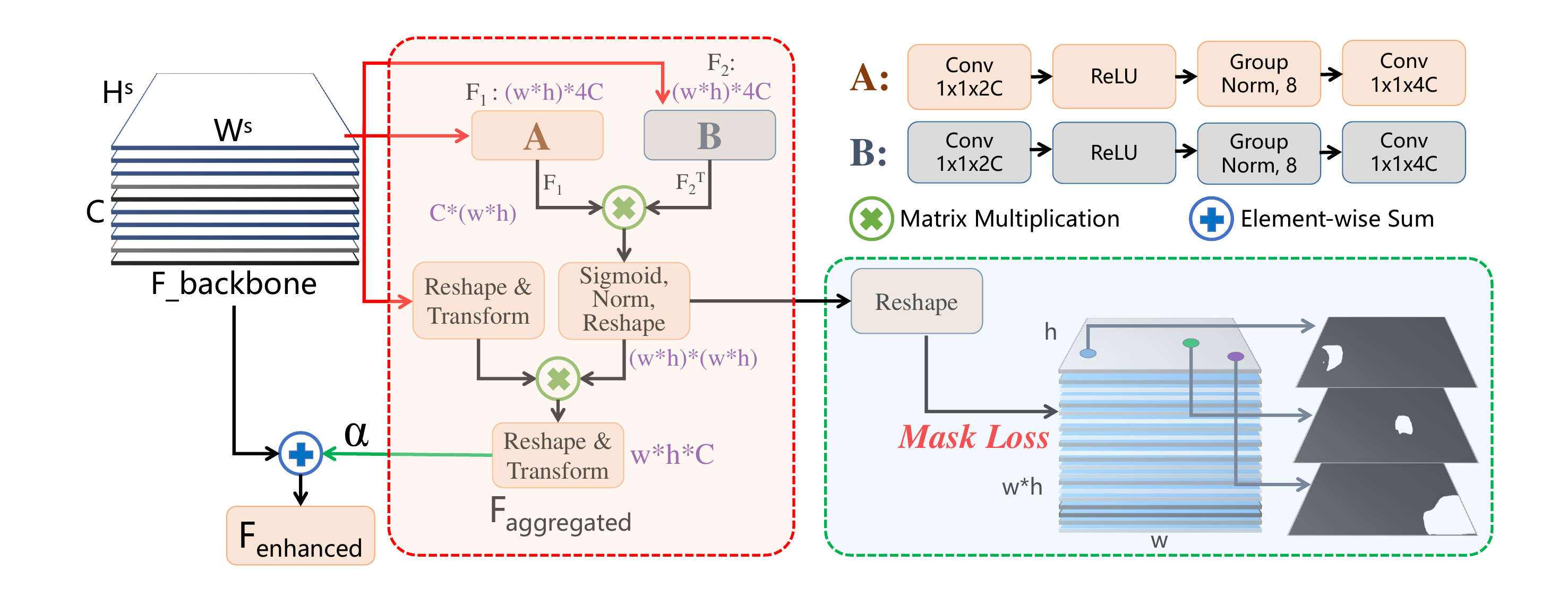}
 	\caption{The proposed IAFA module. The input $\mathbf{F}_\text{backbone}$ will be enhanced by collecting the features from the corresponding instance to generate the output $\mathbf{F}_\text{enhanced}$.  }
 	\label{fig:Feature_aggregation}
 \end{figure}
 
\subsection{Detailed IAFA Structure}
The detailed structure of the proposed IAFA branch is illustrated in \figref{fig:Feature_aggregation}. The input of IAFA is the feature map $\mathbf{F}_\text{backbone} \in \mathbb{R}^{\text{W}^s \times \text{H}^s \times \text{C}}$ from the backbone network and the output of the module is the enhanced feature $\mathbf{F}_\text{enhanced} \in \mathbb{R}^{\text{W}^s \times \text{H}^s \times \text{C}}$ which has the same dimension with the input feature map. The enhanced feature $\mathbf{F}_\text{enhanced}$ can be obtained as 
\begin{equation}
     \mathbf{F}_\text{enhanced}  = \mathbf{F}_\text{backbone} + \alpha * \mathbf{F}_\text{aggregated},
     \label{eq:F_enhanced_computation}
\end{equation}
where $\mathbf{F}_\text{aggregated}$ is the aggregated features from other locations and $\alpha$ is a learn-able parameter initialized with zero to balance the importance of $\mathbf{F}_\text{aggregated}$ and $\mathbf{F}_\text{backbone}$. The building of the  $\mathbf{F}_\text{aggregated}$ has been highlighted with red dotted rectangle in \figref{fig:Feature_aggregation}. If needed, the input $\mathbf{F}_\text{backbone}$ can be downsampled to an appropriate size for saving the GPU memory and upsampled to the same size of $\mathbf{F}_\text{backbone}$ after aggregation. For general representation, we assume the input features size is $\text{w} \times \text{h} \times \text{C}$. First of all, two new feature maps $\{ \mathbf{F}_1, \mathbf{F}_2\} \in \mathbb{R}^{\text{w} \times \text{h} \times 4\text{C}}$ are generated with a series of convolutions operations which are represented with  ``Operation'' \textbf{A} and \textbf{B} in short. Here $\textbf{A}$ and $\textbf{B}$ share the same structure with different parameters. Specifically, both of them contain two $1 \times 1$ convolution layers, one non-linear activation layer (e.g., ReLU) and one group normalization layer. Detailed convolution kernel information and group size information are given in right top of \figref{fig:Feature_aggregation}. Then both of them are reshaped to $ \mathbb{R}^{\text{d} \times 4\text{C}}$, where $\text{d} = \text{w} \times \text{h}$ is the number of the pixels in $\mathbf{F}_1$ or $\mathbf{F}_2$. Assuming the two reshaped tensors are $\mathbf{F}^{'}_1$ and $\mathbf{F}^{'}_2$, then the high-dimension relation map $\mathbf{G}$ 
can be obtained as
\begin{equation}
     \mathbf{G}    = {Norm} \{ {Sigmoid} ( {\mathbf{F}_1 \otimes (\mathbf{F}^{'}_2)^\text{T}}) \},
\end{equation}
where $\otimes$ represent the matrix multiplication, ``\text{Sigmoid}'' represent the $Sigmoid$ function to re-scale the element's value from $(-\infty, +\infty)$ to $(0, +1)$ and \text{Norm} represent the normalization operation along the row dimension. Then we reshape this relationship map $\mathbf{G}$ from $\mathbb{R}^{\text{d} \times \text{d}}$ to $\mathbb{R}^{ \text{W}^{s} \times \text{H}^{s} \times \text{d}}$ and each vector of $\mathbf{G}(i, j)$ gives the relationship of current pixel $(i, j)$ with all other pixels. With the estimated $\mathbf{G}$, $\mathbf{F}_\text{aggregated}$ can be computed as  

\begin{equation}
    \mathbf{F}_\text{aggregated} = \mathbf{G} \otimes \mathcal{F} \{\mathbf{F}^{'}_\text{backbone}\}, 
\end{equation}
here $\mathcal{F} \{ . \}$ operation is used for transforming the downsampled $\mathbf{F}^{'}_\text{backbone}$ from the shape of ${\text{w} \times \text{h} \times \text{C}}$ to the shape of ${\text{d}  \times \text{C}}$. 
Finally, the $\mathbf{F}_\text{aggregated}$ can be upsampled to $\text{W}^{s} \times \text{H}^{s}$, the same size as $\mathbf{F}_\text{backbone}$.

\subsection{Loss Function for Instance Mask}
Three loss functions are used for training the framework which are $L_\text{center-ness}$, $L_\text{reg}$ and  $L_\text{mask}$. Here, we choose the smooth-L1 loss on the 3D BBox's 8 corners for regression loss $L_\text{reg}$. Consequently, the whole loss function is formulated as
\begin{equation}
    \text{L} = \gamma_0 \text{L}_\text{center-ness} + \gamma_1 \text{L}_\text{reg} + \gamma_2 \text{L}_\text{mask},
\end{equation}
where $\gamma_0 $, $\gamma_1$ and $\gamma_3$ are hype-parameters for balancing the contributions of different parts. 
As shown in the green dotted box in \figref{fig:Feature_aggregation}, the loss for mask is only activated sparsely on the center points. Due to the unbalance between the foreground and background pixels, focal loss is also applied here. Similar to \eqref{eq:center-ness-loss}, the $\text{L}_\text{mask}$ is defined with focal loss as
\begin{equation}
   \text{L}_\text{mask}= -\frac{1}{N} \sum^{N}_{j=0} \frac{1}{M_{j}}\sum^{M_{j}}_{i = 0} (1-\hat{y}_{i})^{\alpha} \text{log}(\hat{y}_{i})         
\label{eq:mask_loss}
\end{equation}
where $\hat{y}_{i}$ is the predicted probability that a pixel $i$ belongs to a certain instance $j$, $N$ is the number of instance per batch and $M_j$ is the number of pixels for instance $j$. 

\subsection{Coarse Instance Annotation Generation}

For training the mask attention module, dense pixel-wise instance segmentation annotation is needed. However, for most of the 3D object detection dataset (e.g., KITTI \cite{geiger2012we}), only the 2D/3D bounding boxes are provided and the instance-level segmentation annotation is not provided. In our experiment, we just used the output of the commonly used instance segmentation framework ``Mask R-CNN \cite{he2017mask}'' as the coarse label. Surprisingly, we find that the performance can also be boosted evenly with this kind of noise label.

\section{Experimental Results}
We implement our approach and evaluate it on the public  KITTI \cite{geiger2012we} 3D object detection benchmark. 
\subsection{Dataset and Implementation Details}
\subsubsection{Dataset:} the KIITI data is collected from the real traffic environment in Europe streets. The whole dataset has been divided into training and testing two subsets, which consist of $7,481$ and $7,518$ frames, respectively. Since the ground truth for the testing set is not available, we divide the training data into a training and validation set as in \cite{yan2018second,zhou2018voxelnet}, and obtain $3,712$ data samples for training and $3,769$ data samples for validation to refine our model. On the KITTI benchmark, the objects have been categorized into ``Easy'', ``Moderate'' and ``Hard'' based on their height in the image and occlusion ratio, etc. For each frame, both the camera image and the LiDAR point cloud have been provided, while only RGB image has been used for object detection and the point cloud is only used for visualization purposes.

\subsubsection{Evaluation Metric:} we focus on the evaluation on ``Car'' category because it has been considered most in the previous approaches. In addition, the number of the training data for ``Pedestrain'' and ``Cyclist'' is too small for training a stable model. For evaluation, the average precision (AP) with Intersection over Union (IoU) is used as the metric for evaluation. Specifically, before October 8, 2019, the KITTI test sever used the 11 recall positions for comparison. After that the test sever change the evaluation criterion from 11-points to 40-points because the latter one is proved to be more stable than the former \cite{simonelli2019disentangling}. Therefore, we use the 40-points criterion on the test sever, while we keep the 11-points criterion on validation dataset because most of previous methods only report their performance using 11-points criterion.  

\subsubsection{Implementation Details:} 
for each original image, we pad it symmetrically to $1280 \times 384$ for both training and inference. Before training, these ground truth BBoxes whose depth larger than \SI{50}{\metre} or whose 2D projected center is out of the image range are eliminated and all the rest are used for training our model. Similar to \cite{liu2020smoke}, three types of data-augmentation strategies have been applied here: random horizontal flip, random scale and shift. The scale ratio is set to 9 steps from \numrange{0.6}{1.4}, and the shift ratio is set to 5 steps from \numrange{-0.2}{0.2}. To be clear, the scale and shift augmentation haven't been used for the regression branch because the 3D information is inconsistent after these transformations.   

\textbf{Parameters setting:} for each object, the ``depth'' prediction is defined as $depth = a_0 + b_0 x$, where $a_0$, $b_0$ are two predefined constants and $x$ is the output of the network. Here, we set $a_0 = b_0 = 12.5$ experimentally and re-scale the output $x \in [-1.0, 1.0 ]$. For the focal loss in \eqref{eq:center-ness-loss} and \eqref{eq:mask_loss}, we set $\alpha =2.0$ and $\beta = 4.0$ for all the experiments. The group number for normalization in IAFA module is set to 8. For decreasing the GPU consumption, we set $\text{w} = \frac{1}{2}\text{W}^{s}$ and $\text{h} = \frac{1}{2}\text{H}^{s}$ in the IAFA module.

\textbf{Training:} Adam \cite{kingma2014adam} together with L1 weights regularization is used for optimizing our model. The network is trained with a batch size of 42 on 7 Tesla V100 for 160 epochs. The learning rate is set at $2.5 \times 10^{-4}$ and drops at 80 and 120 epochs by a factor of 10. The total training process requires about 12 hours. During testing, top 100 center points with response above 0.25 are chosen as valid detection. No data augmentation is applied during inference process. 

\subsection{Evaluation on the ``test'' Split}

First of all, we evaluate our methods with other monocular based 3D object detectors on the KITTI testing benchmark. Due the limited space, we only list the results with public publications. For fair comparison, all the numbers are collected directly from the official benchmark website \footnote{\url{http://www.cvlibs.net/datasets/kitti/eval_object.php?obj_benchmark=3d}}. Here, we show the Bird-Eye-View (BEV) and 3D results with threshold of 0.7.

\begin{table*}[ht!]
	\centering
	\resizebox{0.8\textwidth}{!}
	{%
		\begin{tabular}{r| l c c c c c c c}
			\hline
			\multicolumn{1}{c}{\multirow{2}{*}{Methods}} & \multicolumn{1}{c}{\multirow{2}{*}{Modality}} & \multicolumn{3}{c}{\textbf{$\textbf{AP}_\text{3D}$70 (\%)}} & \multicolumn{3}{c}{\textbf{$\textbf{AP}_\text{BEV}$70 (\%)}} & \multicolumn{1}{c}{\multirow{2}{*}{Time (s)}} \\
			\multicolumn{1}{c}{} & \multicolumn{1}{c}{} & \multicolumn{1}{l}{Moderate} & \multicolumn{1}{l}{Easy} & \multicolumn{1}{l}{Hard} & \multicolumn{1}{l}{Moderate} & \multicolumn{1}{l}{Easy} & \multicolumn{1}{l}{Hard} & \multicolumn{1}{c}{}\\ \hline
			FQNet \cite{liu2019deep}  & Mono  & 1.51 & 2.77 & 1.01 &  3.23 & 5.40 & 2.46 & 0.50\\
			GS3D \cite{li2019gs3d} & Mono  & 2.90 & 4.47 & 2.47 &  6.08  & 8.41 & 4.94 & 2.0\\
			MVRA  \cite{choi2019multi} & Mono  & 3.27 & 5.19 & 2.49 &  5.84  &  9.05 & 4.50 & 0.18\\	
			Shift R-CNN \cite{naiden2019shift} &Mono  & 3.87 & 6.88 & 2.83 &  6.82  & 11.84 & 5.27 & 0.25\\
			MonoGRNet \cite{qin2019monogrnet} &Mono &5.74  &9.61 &	4.25 &  11.17   & 18.19 & 8.73 & 0.04\\	
			SMOKE \cite{liu2020smoke} &Mono  & 9.76 & 14.03 &  7.84 & 14.49 & 20.83  &  12.75 & 0.03\\
			MonoPair \cite{chen2020monopair} &Mono  & 9.99 & 13.04 &  8.65 &14.83 & 19.28  &  12.89  & 0.06\\
			RTM3D \cite{li2020rtm3d} & Mono  & 10.34 & 14.41 & 8.77 & 14.20 & 19.17  &  11.99 & 0.05\\
			\hline
			ROI-10D \cite{manhardt2019roi} & $\text{Mono}^{\ast\dagger}$  & 2.02 & 4.32 &	1.46 &  4.91  & 9.78 &	3.74 & 0.20\\
			MonoFENet \cite{bao2019monofenet} & $\text{Mono}^{\ast}$  &5.14 & 8.35 &	4.10 &  11.03  & 17.03 & 9.05 & 0.15\\
			Decoupled-3D \cite{cai2020monocular} &$\text{Mono}^{\ast}$  & 7.02 & 11.08 &  5.63 &14.82 & 23.16  &11.25 & 0.08\\ 
			
			MonoPSR \cite{ku2019monocular} & $\text{Mono}^{\ast}$  & 7.25 & 10.76 &  5.85 & 12.58 & 18.33  & 9.91 & 0.20\\
			AM3D \cite{ma2019accurate} &$\text{Mono}^{\ast}$ & 10.74 & 16.50 &  \B{9.52} & 17.32 & 25.03  &  \B{14.91} & 0.40\\
			RefinedMPL \cite{vianney2019refinedmpl} & $\text{Mono}^{\ast}$   & 11.14& \textbf{18.09} & 8.94 & \B{17.60} &  \textbf{28.08} &  13.95 & 0.15\\ 
			D4LCN \cite{ding2020learning} &$\text{Mono}^{\ast}$   & \B{11.72} & 16.65 &  9.51 & 16.02 & 22.51  &  12.55 & 0.20\\
		 \hline 
		    Baseline \cite{liu2020smoke} &Mono   & 9.76 & 14.03 &  7.84 & 14.49 & 20.83  &  12.75\\
			Proposed Method &Mono & \textbf{12.01} & \B{17.81} & \textbf{10.61}  &  \textbf{17.88} & \B{25.88} & \textbf{15.35} & 0.034\\ 
			Improvement & - & \textbf{\R{+2.25}} &  \textbf{\R{+3.78}} & \textbf{\R{+2.77}}  & \textbf{\R{+3.39}} & \textbf{\R{+5.05}} & \textbf{\R{+2.6}} \\
			\hline  
		\end{tabular}
	}
	\caption{\small Comparison with other public methods on the KITTI testing sever for 3D ``Car'' detection. For the ``direct'' methods, we represent the `` Modality'' with ``Mono'' only. For the other methods, we use $\ast$, $\dagger$ to indicate that the ``depth'' or ``3D model'' have been used by these methods during training or inference procedure. For each column, we have highlighted the top numbers in bold and the second best is shown in blue. The numbers in red represent the absolute improvements compared with the baseline.}
	\label{tab:test_kitti}
\end{table*}

\textbf{Comparison with SOTA methods:} we make our results public on the KITTI benchmark sever and the comparison with other methods are listed in \tabref{tab:test_kitti}. For fair comparison, the monocular-based methods can also be divided into two groups, which are illustrated in the top and middle rows of \tabref{tab:test_kitti}, respectively. The former is called the ``direct''-based method, which only uses a single image during the training and inference. In the latter type, other information such as depth or 3D model is used as an auxiliary during the training or inference. Our proposed method belongs to the former. 

Similar to the official benchmark website, all the results are displayed in ascending order based on the values of ``Moderate'' $\textbf{AP}_\text{3D} 70$. From the table, we can find that the proposed method outperforms all the ``direct''-based method with a big margin among all the three categories. For example,  for ``Easy'' $\textbf{AP}_\text{BEV} 70$, our method achieved $5.05$ points improvements than the best method of SMOKE \cite{liu2020smoke}. The minimum improvement happens on ``Moderate'' $\textbf{AP}_\text{3D} 70$, even so, we also obtained $1.67$ points of improvements than RTM3D \cite{li2020rtm3d}.   

Based on the evaluation criterion defined by KITTI (ranking based on values of ``Moderate'' $\textbf{AP}_\text{3D} 70$ ), our method achieved the first place among all the monocular-based 3D object detectors \footnote{Only these methods with publications have been listed for comparison here.} up to the submission of this manuscript (Jul. 8, 2020), including these models trained with depth or 3D models. Specifically, the proposed method achieved four first places, two second places among all the six sub-items. The run time of different methods is also provided in the last column of \tabref{tab:test_kitti}. Compared with other methods, we also show superiority on efficiency. By using the DAL34 as the backbone network, our methods can achieve 29 fps on Tesla V-100 with a resolution of $384 \times 1280$.  

\textbf{Comparison with baseline:} from the table, we also can find that the proposed method significantly boosts the baseline method on both the BEV and 3D evaluation among all the six sub-items. Especially, for ``Easy'' category, we have achieved ${3.78}$ and ${5.05}$ points improvements for $\textbf{AP}_\text{3D}$ and $\textbf{AP}_\text{BEV}$ respectively. For the other four sub-items, the proposed method achieves more ${2.0}$ points improvement. The minimal improvement we have achieved is for ``Moderate'' $\textbf{AP}_\text{3D} \textbf{70}$, while it also provides ${2.25}$ points of improvement.

\begin{table*}[ht!]
	\centering
	\resizebox{0.65\textwidth}{!}
	{%
		\begin{tabular}{r| l ccc ccc}
			\hline
			\multicolumn{1}{c }{\multirow{2}{*}{Methods}}& \multicolumn{1}{c}{\multirow{2}{*}{Modality}} & \multicolumn{3}{c}{\textbf{$\textbf{AP}_\text{3D}$70 (\%)}} & \multicolumn{3}{c}{\textbf{$\textbf{AP}_\text{BEV}$70 (\%)}}\\
			\multicolumn{1}{c}{} & \multicolumn{1}{c}{} & \multicolumn{1}{l}{Mod } & \multicolumn{1}{l}{Easy} & \multicolumn{1}{l}{Hard} & \multicolumn{1}{l}{Mod} & \multicolumn{1}{l}{Easy} & \multicolumn{1}{l}{Hard}\\ \hline
			CenterNet \cite{zhou2019objects}  & Mono    & 1.06  & 0.86 & 0.66 &  3.23 & 4.46 & 3.53\\    
			Mono3D \cite{chen2016monocular}  & Mono  & 2.31 & 2.53 &	2.31  &  5.19 & 5.22 &	4.13\\   
			OFTNet  \cite{roddick2018orthographic} & Mono  & 3.27 &  4.07 & 3.29 &  8.79   & 11.06 & 8.91\\
			GS3D \cite{li2019gs3d} & Mono  & 10.51 & 11.63 & 10.51 &  - & - & - \\  
			MonoGRNet \cite{qin2019monogrnet} &Mono  & 10.19  & 13.88 &	7.62  &  -   & - & -\\ 
			ROI-10D \cite{manhardt2019roi} & Mono  & 6.63 &  9.61 &	6.29 &  9.91  & 14.50 &	 8.73 \\  
			MonoDIS \cite{simonelli2019disentangling} &Mono  & \B{14.98} & 18.05 & \B{13.42} & 18.43   &  \B{24.26} & 16.95\\  
			M3D-RPN \cite{brazil2019m3d} &Mono & \textbf{16.48} & \textbf{20.40} & 13.34 &  \B{21.15}   & \textbf{26.86} & \B{17.14}\\	 
				   
			\hline 
			Baseline \cite{liu2020smoke} &Mono & 12.85 & 14.76  & 11.50 & 15.61     & 19.99 & 15.28 \\
			Proposed &Mono & 14.96 & \B{18.95} & \textbf{14.84}  &  \B{19.60} & 22.75 & \textbf{19.21} \\  
			Improvements&  & \textbf{\R{+2.11}} & \textbf{\R{+4.19}} & \textbf{\R{+3.34}}  & \textbf{\R{+3.998}} & \textbf{\R{+2.76}} & \textbf{\R{+3.94}} \\ \hline 
		\end{tabular}   
	}
	\caption{\small Comparison with other public methods on the KITTI ``val'' split for 3D ``Car'' detection, where ``-`` represent the values are not provided in their papers. For easy understanding, we have highlighted the top number in bold for each column and the second best is shown in blue. The numbers in red represent the absolute improvements compared with the baseline.}
	\label{tab:val_kitti}
\end{table*}
\subsection{Evaluation on ``validation'' Split}
We also evaluate our proposed method on the validation split. The detailed comparison is given in \tabref{tab:val_kitti}.  As mentioned in \cite{wang2019pseudo}, the 200 training images of KITTI stereo 2015 overlap with the validation images of KITTI object detection. That is to say, some LiDAR point cloud in the object detection validation split has been used for training the depth estimation networks. That is the reason why some 3D object detectors (with depth for training) achieved very good performances while obtained unsatisfactory results on the test dataset. Therefore, we only list the ``direct''-based methods for comparison here. Compared with the baseline method, the proposed method achieves significant improvements among all the six sub-items. Especially, we achieve more than $\B{3.0}$ points improvement in four items and the improvements for all the items are above $\B{2.0}$ points. Comparison with other methods, we achieve \R{2} first places, \R{2} second places and \R{2} third places among all the 6 sub-items. 


\subsection{Ablation Studies}
In addition, we also have designed a set of ablation experiments to verify the effectiveness of each module of our proposed method. 
\begin{table*}[ht!]
	\centering
	\resizebox{0.5\textwidth}{!}
	{%
		\begin{tabular}{r  ccc ccc}
			\hline
			\multicolumn{1}{c}{\multirow{2}{*}{Methods}}&   \multicolumn{3}{c}{\textbf{$\textbf{AP}_\text{3D}$70 (\%)}} & \multicolumn{3}{c}{\textbf{$\textbf{AP}_\text{BEV}$70 (\%)}} \\
			\multicolumn{1}{c}{}     & \multicolumn{1}{l}{Mod } & \multicolumn{1}{l}{Easy} & \multicolumn{1}{l}{Hard} & \multicolumn{1}{l}{Mod}  & \multicolumn{1}{l}{Easy} & \multicolumn{1}{l}{Hard}\\ \hline
			Baseline         & 12.85 & 14.76 & 11.50 & 15.61 & 19.99 & 15.28 \\
			w/o supervision & 12.98 & 14.59 & 11.76 &  15.79 &20.12 & 14.98\\  
			w  supervision   & 14.96 & 18.95 & 14.84 & 19.60 & 22.75 & 19.21\\    
		    \hline 
		\end{tabular}   
	}
	\caption{\normalfont Ablation studies on the KITTI ``val'' split for 3D ``Car'' detection with/without instance mask supervision. From the table, we can easily find that the performances have been largely improved by adding the mask supervision signal.}
	\label{tab:mask_influence}
\end{table*}
\subsubsection{Supervision of the instance mask:} 
\begin{figure}[ht!]
 	\centering
 	\includegraphics[width=0.995\textwidth]{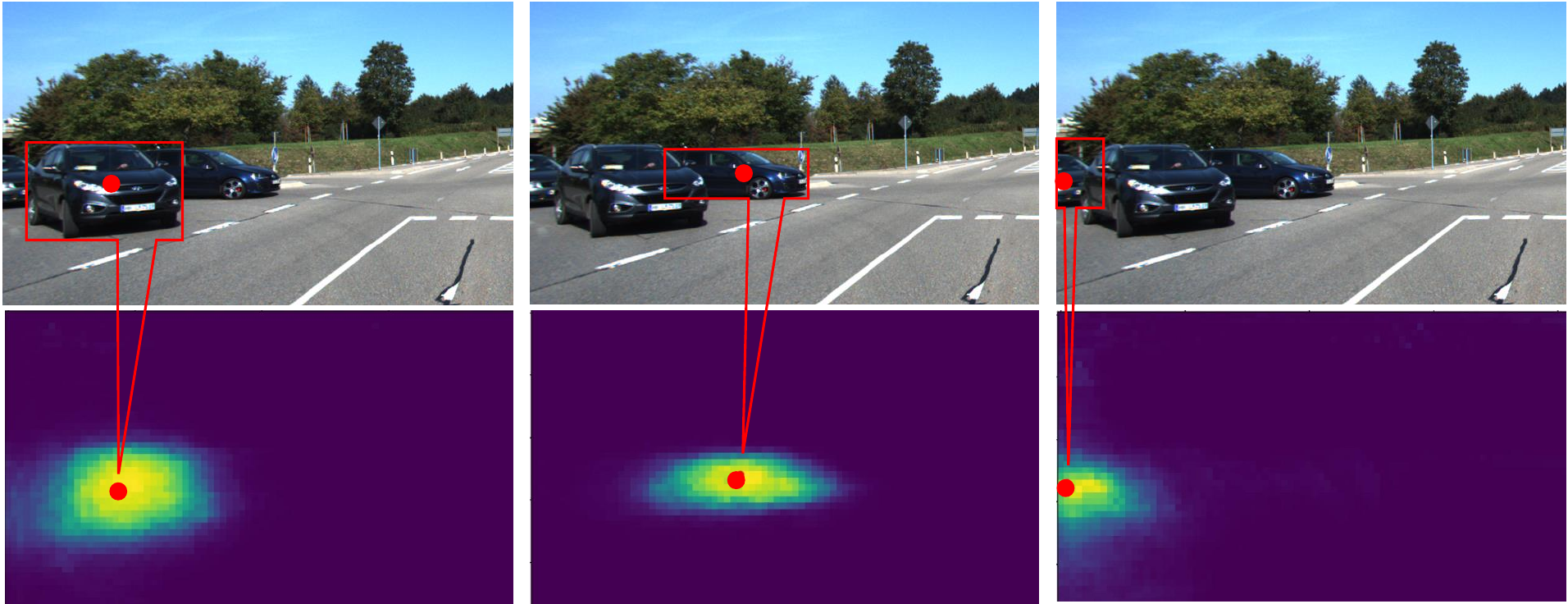}
 	\caption{An example of feature map for IAFA module. Different brightness reveals different importance related to the target point (red dot). }
 	\label{fig:Example_of_AttentionMap}
 \end{figure}

self-attention strategy is commonly used for in semantic segmentation \cite{liu2019deep}, \cite{fu2019dual} and object detection \cite{gu2018learning} etc. To highlight the influence of the supervision of the instance mask, we compare the performance of the proposed module with and without the supervision signal. From the table, we can easily found that the supervision signal is particularly useful for training IAFA module. Without the supervision, the detection performance nearly unchanged. The positive effect of the instance supervision signal is obvious. Furthermore, we also illustrated some examples of the learned attention maps in \figref{fig:Example_of_AttentionMap}, where the bottom sub-figures are the corresponding attention maps for the three instances respectively. From the figure, we can see that the maximum value is at the center of object and it decreases gradually with the increasing of its distance to the object center. 
 \subsection{Qualitative Results}
\begin{figure}[ht!]
 	\centering
 	\includegraphics[width=0.995\textwidth]{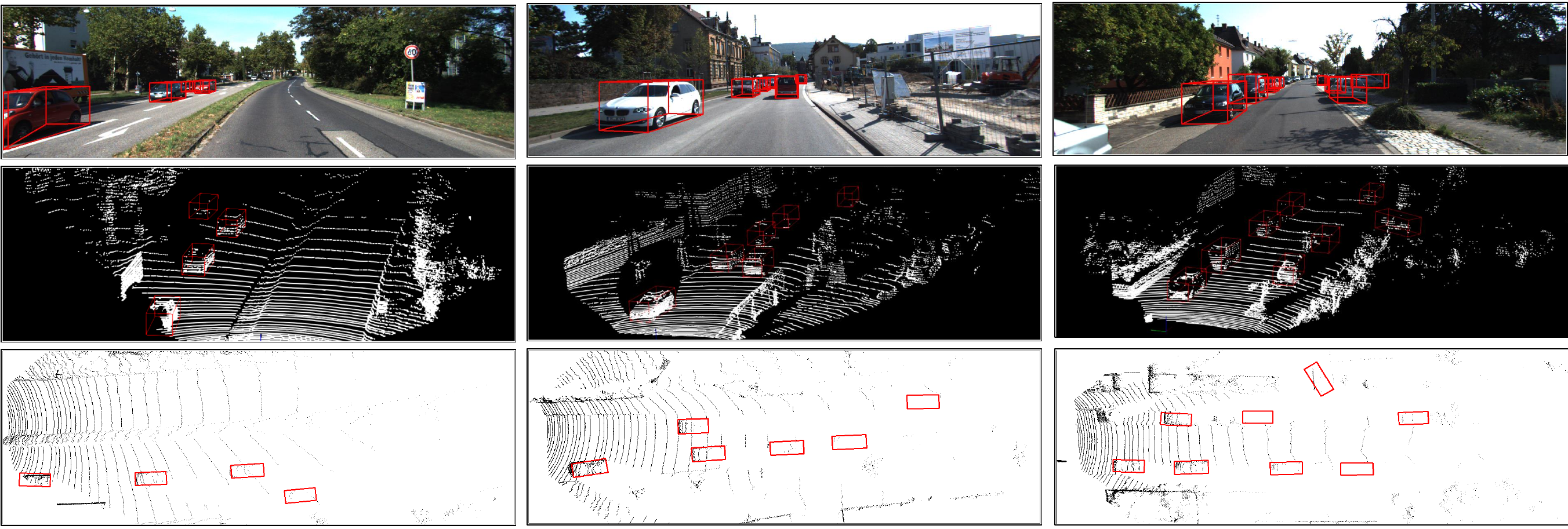}
 	\caption{Three examples of 3D detection results. The ``top'', ``Middle'' and ``Bottom`` are results are drawn in RGB image, 3D Point cloud and BEV-view respectively. The point cloud is only used for visualization purposes here. }
 	\label{fig:qualitative_results_1}
 \end{figure}

 Some qualitative detection results on the test split are displayed in \figref{fig:qualitative_results_1}. For better visualization and comparison, we also draw the 3D BBoxes in point cloud and BEV-view images. The results clearly demonstrate that the proposed framework can recover objects' 3D information accurately. 


\section{Conclusions and Future Works}
In this paper, we have proposed a simple but effective instance-aware feature aggregation (IAFA) module to collect all the related information for the task of single image-based 3D object detection. The proposed module is an easily implemented plug-and-play module that can be incorporated into any one-stage object detection framework. In addition, we find out that the IAFA module can achieve satisfactory performance even though the coarsely annotated instance masks are used as supervision signals.

In the future, we plan to implement the proposed framework for real-world AD applications. Our proposed framework can also be extended to a multi-camera configuration to handle the detection from \SI{360}{\degree}- viewpoints. In addition, extending the detector to multi-frame is also an interesting direction, which can boost the detection performances of distant instances.

\bibliographystyle{unsrt}
\bibliography{egbib}

\end{document}